6th International Conference on AI in Computational Linguistics

# Exploring Retrieval Augmented Generation in Arabic

Samhaa R. El-Beltagy* and Mohamed A. Abdallah

*Newgiza University, Newgiza, km 22 Cairo-Alex Desert Rd, Cairo, Egypt*

**Abstract**

Recently, Retrieval Augmented Generation (RAG) has emerged as a powerful technique in natural language processing, combining the strengths of retrieval-based and generation-based models to enhance text generation tasks. However, the application of RAG in Arabic, a language with unique characteristics and resource constraints, remains underexplored. This paper presents a comprehensive case study on the implementation and evaluation of RAG for Arabic text. The work focuses on exploring various semantic embedding models in the retrieval stage and several LLMs in the generation stage, in order to investigate what works and what doesn't in the context of Arabic. The work also touches upon the issue of variations between document dialect and query dialect in the retrieval stage. Results show that existing semantic embedding models and LLMs can be effectively employed to build Arabic RAG pipelines.

## 1. Introduction

Retrieval-Augmented Generation (RAG) models have recently emerged as powerful tools that can both enhance and capitalize on the capabilities of generative systems through integration with external knowledge source [1]. The advantage of using a RAG model is that it leverages the power of large language models (LLMs) to generate responses based on documents that these LLMs might not have seen before. In specific domains, this means getting high-quality and accurate answers to queries to which an LLM might not have an answer. In most scenarios, the use of a RAG model also reduces LLM hallucinations.

---

* Samhaa R. El-Beltagy.
  *E-mail address:* samhaa@computer.org

While extensive research has been conducted on the application and effectiveness of RAG in English, this has not been the case for almost all other languages, and even less so for Arabic. According to Wikipedia, the Arabic language is spoken by approximately 422 million people, making it one of the most widely used languages in the world[1]. As a language, Arabic has unique linguistic characteristics, and its automatic processing is often complicated by a diverse set of dialects used across the many countries in which it is the official language [2]. Not only do these dialects vary significantly from one region to another, but they are also quite distant from Modern Standard Arabic (MSA), which is the formal written version of Arabic.

A typical RAG system is composed of various components, as shown in Fig 1. One of the most important components is the retriever, which is the entity responsible for retrieving pieces of text that act as the context from which the generator can formulate a final response for the user. Retrieval is a crucial step because it allows the model to augment its pre-existing knowledge with specific, up-to-date information, leading to outputs that are not only well-informed and accurate, but also tailored to the specifics of the user query. Failure to retrieve correct pieces of text to pass on to the generator means that the whole system will not work as expected. Since most generators have context limitations, it is best if the retriever's top results are the ones from which an answer can be extracted. The use of semantically rich embeddings has been shown to be the best way to approach this task. However, given the fact that the recent most powerful LLMs, which are often used to generate embeddings, have been predominantly trained on English documents and that the extent of support for Arabic in multilingual models is not known, one of the main goals of this study is to investigate the retrieval aspect of RAG—specifically, how various embedding models perform in the context of Arabic. This includes investigating which embeddings are most effective for capturing the semantic nuances necessary for accurate retrieval and whether the retrieval process itself is impacted by the linguistic variations inherent to Arabic dialects. The second aim of the work presented in this paper is to also do some preliminary analysis to investigate which LLMs work best as generators in Arabic with focus on open-source models. The study is by no means comprehensive, but opens the door for further investigations in this area. All code and data files related to this investigation are publicly available [2].

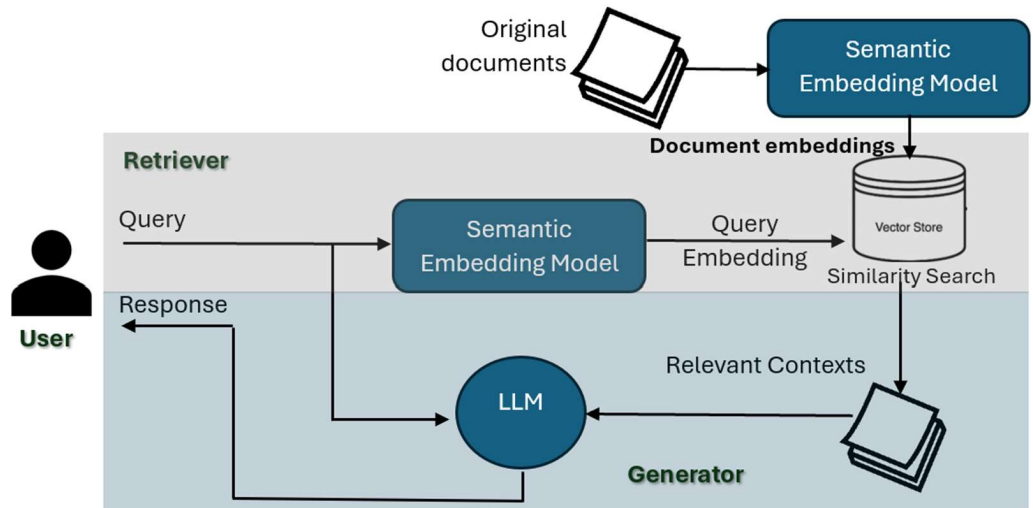

Fig. 1. A typical RAG System Architecture

---

[1] Wikipedia: List of countries and territories where Arabic is an official language (https://shorturl.at/WpcYU)
[2] https://github.com/SElBeltagy/ArRagExperiments

The rest of this paper is organized as follows: Section 2 provides a short overview of related work. Section 3 describes the methodology used in this study as well as the experimental setup. Section 4 outlines the carried out experiments and their results, while section 5 concludes this paper with a summary of key insights gained and suggestions for future research directions.

## 2. Related Work

The concept of Retrieval-Augmented Generation (RAG) has gained significant attention as a hybrid approach that integrates information retrieval with neural language generation. In 2020, Lewis et al. [3] introduced the RAG framework to address the then-limited capabilities of pre-trained language models and to enhance a model's ability to generate informed and contextually relevant responses. Despite the major improvements in language model capabilities since then, RAG remains highly relevant for generating accurate and contextually enriched responses by dynamically accessing and synthesizing information from external sources [1] [4].

While the literature is rich with issues related to the application of RAG on English documents, the application of RAG in languages other than English has been less explored. However, recent studies have begun to address this gap. For example, the study presented in [5], discusses the application of Retrieval-Augmented Generation (RAG) in multilingual settings, specifically focusing on enhancing the performance of RAG models when working with non-English languages. The work emphasizes the need for strong retrievers and generators and highlights the importance of task-specific prompt engineering to generate responses in a user's language. The paper suggests that while the multilingual RAG models show promise, they face challenges with code-switching, fluency errors, and the relevance of retrieved documents.

The work presented in [6] specifically addresses the effectiveness of multilingual semantic embedding models for Arabic text retrieval, making it highly relevant to the study discussed here, as both explore the retrieval aspect of Arabic RAG models. The experiments presented in that work were carried out using the publicly available ARCD (Arabic Reading Comprehension Dataset) [7]. These experiments involved assessing the performance of several advanced multilingual semantic embedding models in retrieving text passages relevant to a query using the average Recall@k metric. The authors did not employ a vector database and chose to directly use cosine similarity for matching query embeddings against document embeddings. While this can slow down the matching process in a real-life setting, it should have little or no impact on the research findings presented in the paper. The embedding models investigated by this study were OpenAI's Ada [8], Google's Language-agnostic BERT Sentence Embedding (LaBSE) [9], Cohere [10], Mpnet [11], HuggingFace's DistillBert versions one and two [12], Meta's SONAR (Language-Agnostic Representations)[13], and Microsoft's E5 embedding models[14]. The study identified Microsoft's E5 large sentence embedding model as the top performer, significantly outperforming other models tested.

While the work presented here also uses the ARCD dataset [7] and experiments with OpenAI's Ada model [8], Cohere [10] and Microsoft's E5 embedding models[14], the work goes further by extending the experiments to other embedding models, using a second dataset for experimentation, examining the impact of using dialectical queries on the performance of embeddings when carrying our retrieval, and investigating the impact of attempting to eliminate ambiguity in ARCD queries. Furthermore, the work presented herein investigates several known LLMs as generators to present an exploration of a complete RAG pipeline.

## 3. Methodology and Experimental Setup

One of the main aims of this work is to assess the performance of various multilingual semantic embedding models in the context of Arabic text retrieval, and to test the resilience of top performing models to a query dialect different than that of input documents. The work also aims to evaluate the performance of multilingual Large Language Models (LLMs) for the generation task. To accomplish these goals, the authors set out to implement the entire pipeline presented in Fig 2 over 2 stages. In the first stage, experiments are carried out to identify the best semantic model to use, and in the second stage experiments are conducted to evaluate the performance of various LLMs as generators using the best performing semantic model in stage 1 in the retrieval process.

Details of the used datasets, semantic embedding models, vector database, and LLMs used as generators are provided in the next subsections.

*3.1. Used Datasets*

In this work, two different datasets were used for experimentation. The first is the Arabic EduText Secondary School dataset which was compiled by the authors while the second is the ARCD (Arabic Reading Comprehension Dataset) [7]. Each of the datasets is briefly described below.

*3.1.1. The Arabic EduText Secondary School Dataset  (Ar_EduText)*
The goal of creating this dataset was to facilitate the testing of multiple embedding and generation models within a manageable scope. The dataset was compiled by randomly selecting six freestyle reading passages from high school Arabic textbooks which are written in MSA. Each passage was input to OpenAI's GPT 4o[3] model using the prompt: "You are an expert in Arabic. Given the following text (a paragraph), create five or six different Arabic questions." The generated questions were manually reviewed, and depending on their suitability, they were either left as is, edited, or rejected. This process yielded a set of 158 distinct questions, each linked to the text segment from which it was generated. To provide answers for the questions as part of the dataset, each segment from which a question originated was submitted back to OpenAI's GPT-3.5 Turbo model[4], along with the question and the prompt: "Given the following context (segment text) and the following question (edited question), provide a concise answer." The answers were also manually reviewed, and both the automatically generated and edited versions were retained. A final step was carried out to generate an Egyptian dialect version of the questions. The objective of this step was to generate data that can be used to test the ability of semantic representation models to capture semantic similarities across different dialectal representations. To generate the Egyptian Arabic version of the questions, the original question and the following prompt were passed to the GPT-3.5 Turbo model: "You are fluent in Arabic and its variations. Rewrite the following question in the Egyptian Arabic dialect: [question]." The outputs were manually reviewed and edited by the authors, who are fluent in Egyptian Arabic. The generated Egyptian dialect questions often suffered from structural issues, and frequently included Modern Standard Arabic (MSA) terms instead of Egyptian Arabic terms (e.g., 'حرائق' instead of 'حرايق ' and 'مياه' instead of 'ميه'). These issues were resolved after editing 75.3% of the generated questions, and both the auto-generated and edited versions were retained. The file containing all segments, their associated questions, their auto-generated answers, their manually revised answers, their auto-generated Egyptian dialect versions of the questions, and their corresponding corrections were then saved and are available for download from the project's GitHub repository[5].

 This dataset is intentionally compact and was created as such to enable detailed revisions of answers, and the generation and refinement of questions, particularly those articulated in the Egyptian dialect.

*3.1.2. The Arabic Reading Comprehension Dataset (ARCD)*
The Arabic Reading Comprehension Dataset (ARCD)[7] is composed of 1,395 questions and their answers crowd-sourced from 155 Wikipedia articles spanning diverse domains. Each entry in the dataset is also associated with a paragraph from which an answer can be extracted. In total, there are 460 unique paragraphs in the dataset. The dataset was specifically developed to address the scarcity of Arabic question answering (QA) datasets.

Upon reviewing the data, the authors observed that many of the questions could only be fully understood when considered in the context of the immediately preceding question, as shown in Table 1. A query like the one in the second row, for example, cannot yield any meaningful results, regardless of the expressiveness of the semantic embedding model employed. This problem is not specific to the Arabic language and can occur across different datasets spanning various languages when contextual dependencies between questions influence their interpretation. To eliminate the impact of these dependencies on the results of the carried-out experiments, the authors of this work attempted to disambiguate questions with dependencies. Towards this end,  all questions were input to the GPT 3.5 Turbo model along with the prompt found in Appendix A. Disambiguating the questions using a straight forward prompt with no examples, often produced unexpected results as well some English responses which is why the long prompt shown in the appendix was used. The disambiguator always used the version of the preceding question that was already disambiguated as context for the question being disambiguated. When revising the output of this process it was observed that as a side effect, typos were corrected, and some questions were occasionally rephrased. It was also observed that in a few cases, there were errors in the automatically disambiguated questions. Since the goal was not to find the best way to automatically disambiguate questions, but rather to examine the impact of disambiguation on the retrieval accuracy and hence gain a better understanding of the ability of the semantic embedding model being used, all disambiguated questions were manually revised and edited. Questions that were changed by GPT3.5 for no apparent reason, were restored to their original form. If a question could not be disambiguated in light of its disambiguated preceding question, it was left as is. In the retrieval experimentation section, results are reported on the original questions, the automatically disambiguated questions, and the manually edited disambiguated questions.

---

[3] https://platform.openai.com/docs/models/gpt-4o
[4] https://platform.openai.com/docs/models/gpt-3-5-turbo
[5] https://github.com/SElBeltagy/ArRagExperiments

Table 1. An example of interdependency between questions

| Question | Translation | Disambiguated Translation |
|---|---|---|
| من هو نزار القباني؟ | Who is Nizar Al-Qabbani? | No change |
| متى ولد ؟ | When was he born? | When was Nizar Al-Qabbani born? |

*3.2. The Used Semantic Embedding Models*

Semantic embeddings or text embeddings offer a way of representing text where a word, phrase, sentence, paragraph, or an entire document is represented as a dense vector of real numbers that captures the meaning of what it represents. Since these representations exist in a high-dimensional vector space, distance metrics such as cosine similarity can be applied to evaluate how similar or distant certain pieces of text are from one another. This method can thus facilitate a deeper understanding of textual relationships by quantifying semantic similarities and differences. The concept of semantic embeddings is not new. One of the earliest models in this field is Latent Semantic Analysis (LSA), which was developed in the late 1980s and early 1990s [15, 16], but embeddings gained popularity and widespread use after the introduction of Word2Vec models[17]. The introduction of contextualized embeddings [18, 19] transformed and revolutionized the way in which text is handled and processed.

Table 2. Summary of used embedding models.

| Embedding Name | Model | Dimension | Free |
|---|---|---|---|
| AraBERT | aubmindlab/bert-base-arabertv02 | 768 | Yes |
| AraVec (Wikipedia – CBOW) | wikipedia_cbow_300 | 300 | Yes |
| AraVec (Wikipedia – SG) | wikipedia_sg_300 | 300 | Yes |
| BGE | bge-m3 | 1024 | Yes |
| Cohere 1 | embed-multilingual-v3.0 | 1024 | No |
| Cohere 2 | multilingual-22-12 | 768 | No |
| E5-Large | multilingual-e5-large | 1024 | Yes |
| E5-Small | multilingual-e5-small | 384 | Yes |
| JAIS (13B Q) | core42_jais-13b-bnb-4bit | 5120 | Yes (for research) |
| Ollama | nomic-embed-text | 768 | Yes |
| OpenAI | text-embedding-ada-002 | 1536 | No |

In this work, AraVec[20] which is the name given to a series of different models trained separately on Twitter, and Arabic Wikipedia using Word2Vec CBOW and Skip-gram architectures was primarily used as a baseline and since the datasets used in this paper mostly resemble Wikipedia data, AraVec's Wikipedia models were the ones experimented with. Another baseline model that this work explores is AraBERT[21]. AraBERT is a BERT model pre-trained on a vast corpus of Arabic data collected from various sources [21]. At the time of its introduction, it achieved state of the art performance on a number of downstream tasks including question answering. State of the art semantic models that this work experiments with are those provided by OpenAI [8], Cohere [10] Microsoft's E5 [14], Ollama [22], JAIS [23], and BAAI general embedding (BGE) [24]. The reason we wanted to experiment with JAIS is that it has been launched as the "state-of-the-art Arabic-centric foundation and instruction-tuned open generative large language model" [23]. Unfortunately, its smallest full model is quite heavy, and we were unable to run it using available computational resources so in order to obtain results, we used its quantized version. As for BGE, even though it was designed primarily for the Chinese language, its multilingual version was trained on Arabic documents and the aim of this investigation was to capture the extent to which this model can handle Arabic. A summary of used models is shown in Table 2.

*3.3. The Vector Database*

The vector store that we chose to use is Chroma DB[6] which is a free, open source, easy to use database that can be efficiently employed to store and retrieve embedding vectors. While Chroma might not be the best choice for very large datasets, given the size of the datasets we experimented with, it is ideal.

*3.4. The Generators*

In the context of RAG models, the generator is the component that takes as input the user-entered query and the pieces of text likely to contain an answer (context), and generates an informative and concise response for the question from the context. Typically, the generator is a large language model, and the query and context are presented to it in the form of a prompt with the following structure: "Use the given context to answer the given question. Be as concise as possible. Context: {context}, Question: {question}." While this is the general structure of the prompt, it is often expanded based on the particular nature and requirements of the RAG being developed.

In this work, we experimented with 5 different LLMs as generators. Those are: OpenAI's GPT3.5 Turbo, Mistral 7B[25], Llama 3[26], Mixtral [27], and JAIS [23]. To evaluate the various models, the Precision, Recall, and F1 Score metrics were borrowed from the information retrieval and question answering (QA) domains. In the context of QA, these metrics are calculated based on the overlap of tokens between the system-generated answer and the provided gold standard answer. Another used metric is the BLEU score (Bilingual Evaluation Understudy) which is borrowed from the field of machine translation. In the context of QA, BLEU measures how well a system-generated answer matches a set of reference answers by calculating the precision of n-grams (sequences of n words) in the generated answer against those in the reference answer. The main issue with metrics like BLEU, precision, recall, and F-score is that they primarily focus on exact matches and surface-level features, often failing to capture semantic similarity. To overcome this limitation, the cosine similarity metric was also used to compare the embeddings of the system generated response to the embeddings of the gold standard response.

**4. Experiments and Results**

*4.1. Retriever Related Experiments*

*4.1.1. Experiment 1: Investigating the impact of different semantic embedding models on retrieval*

The goal of this experiment was to assess which semantic embedding models have the highest retrieval rates. In the datasets used, each question was associated with a segment from which an answer could be retrieved. The evaluation focused on the effectiveness of various models in accurately identifying and extracting relevant text segments based on the input queries. To this end, embeddings for segments were generated using the semantic embedding model being tested and stored in Chroma. Query embeddings were then generated using the same model and used to retrieve the top 5 matches from Chroma. Average recall@k (equation 1) was employed as one of two metrics to quantify how many of the correct answers appeared within the top 'k' results provided by each model, thereby determining the models' ability to retrieve necessary information from the dataset. The second employed metric is Mean Reciprocal Rank (MRR) which is a statistical measure used to evaluate the performance of query response systems. MRR is calculated as the average of the reciprocal ranks of results for a set of queries as shown in equation 2. Essentially, MRR calculates the mean of these reciprocal ranks over all queries tested. Higher MRR values indicate that the correct answers tend to appear earlier in the list of responses, which is desirable as the context being passed to a generator is usually limited.

---

[6]https://www.trychroma.com

$$recall@k = \frac{\text{Number of Relevant Items in Top K}}{\text{Total number of relevant items}} \qquad (1)$$

$$MRR = \frac{1}{|Q|}\sum_{i=1}^{|Q|}\frac{1}{rank_i} \qquad (2)$$

where |Q| is the number of queries and $rank_i$ is the rank position of the first relevant segment for the i-th query.

This approach allowed for a direct comparison of the models' performance in real-world retrieval tasks. For the Ar_EduText dataset, only 19 segment embeddings were stored in Chroma, while 460 segment embeddings were stored for the ARCD dataset. The results of this experiment are shown in Table 3, and 4 and in Fig. 2. While all twelve semantic models were used with the first dataset, only the top performing seven were used in the second.

Table 3. Recall @K (k=1, k=3, and k=5) and MRR for the Ar_EduText Dataset (sorted by MRR)

| Embeddings Name | k=1 | k=3 | k=5 | MRR |
|---|---|---|---|---|
| E5-Large | **0.88** | **0.994** | **0.994** | **0.934** |
| E5-Small | 0.867 | 0.981 | 0.987 | 0.92 |
| BGE | 0.861 | 0.956 | **0.994** | 0.9 |
| OpenAI | 0.823 | 0.968 | **0.994** | 0.895 |
| AraVec (Wikipedia – SG) | 0.665 | 0.892 | 0.924 | 0.778 |
| Cohere 2 | 0.62 | 0.829 | 0.88 | 0.73 |
| AraBert | 0.595 | 0.816 | 0.899 | 0.712 |
| AraVec (Wikipedia – CBOW) | 0.595 | 0.823 | 0.911 | 0.718 |
| Ollama | 0.165 | 0.316 | 0.386 | 0.246 |
| JAIS (13B Q) | 0.063 | 0.253 | 0.418 | 0.181 |
| Cohere 1 | 0.032 | 0.133 | 0.285 | 0.111 |

Table 4. Recall @K (k=1, k=3, and k=5) and MRR for the ARCD Dataset (sorted by MRR)

| Embeddings Name | k=1 | k=3 | k=5 | MRR |
|---|---|---|---|---|
| E5-Large | **0.686** | **0.896** | **0.927** | **0.79** |
| BGE | 0.627 | 0.872 | 0.909 | 0.748 |
| E5-Small | 0.632 | 0.861 | 0.893 | 0.742 |
| OpenAI | 0.515 | 0.736 | 0.79 | 0.627 |
| AraVec (Wikipedia – SG) | 0.362 | 0.564 | 0.627 | 0.468 |
| AraBert | 0.318 | 0.514 | 0.574 | 0.418 |
| AraVec (Wikipedia – CBOW) | 0.234 | 0.39 | 0.466 | 0.321 |

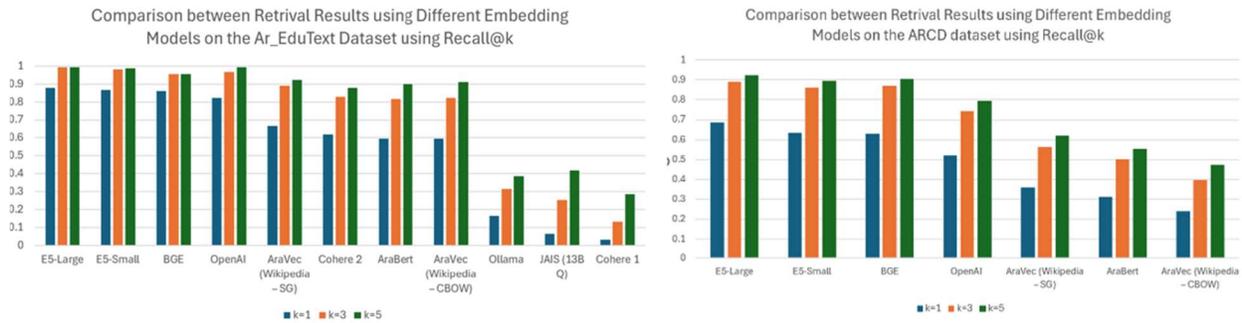

Fig. 2. (a) Retrieval results on the Ar_EduText Dataset; (b) Retrieval results on the ARCD Dataset.

As can be seen from the results, the best performing model is Microsoft's E5 [14] with both its large and small variations ranking at 1 and 2 respectively at k=1 on both used datasets. The E5 large model ties with OpenAI at k@5 for the first dataset, but performs much better on the second. This is very much in line with the results presented in [6]. In the first experiment, Aravec's Wikipedia skip-gram model does surprisingly well, outperforming both Cohere models as well as Ollama and JAIS. That the JAIS quantized model came close to the bottom in terms of performance was quite unexpected, but since we did not experiment on the full model, no concrete conclusion can be reached in terms of JAIS except to say that other models were easier to use and deploy and that the used quantized model is not likely to perform well on Arabic datasets. The BGE model also performed quite well on both datasets ranking at number 1 with respect to recall@5 along with E5 large and OpenAI on the first dataset, and at number 2 on the same metric as well as with respect to MRR on the second dataset, making it a good contender as a semantic model when dealing with Arabic text. It was observed that the results in both experiments were more or less consistent in terms of the ranking of the performance of the various models.

### 4.1.2. Experiment 2: Investigating the impact of using a different dialect on retrieval results

As stated earlier, dialects pose a serious challenge when dealing with Arabic text. It is often the case that a user of a RAG system might choose to interact with it using their local dialect. If the user's query cannot be matched to the text segment from which a response can be generated, no appropriate response will be generated. To evaluate the impact of dialect-specific queries on system performance where the text segments are represented in one variant (MSA here) and the query in another, the experiment described in the previous section was repeated using the Egyptian Arabic version of the questions to create query embeddings. In this experiment, only the Ar_EduText dataset was used. The results are shown in table 5.

Table 5. Recall @K (k=1, k=3, and k=5) and MRR for the Ar_EduText dataset with Egyptian Arabic questions

| Embeddings Name | k=1 | k=3 | k=5 | MRR |
| --- | --- | --- | --- | --- |
| E5-Large | 0.81 | **0.975** | **0.994** | **0.891** |
| BGE | 0.804 | **0.975** | **0.994** | 0.886 |
| E5-Small | **0.817** | 0.937 | 0.962 | 0.879 |
| OpenAI | 0.778 | 0.93 | 0.968 | 0.855 |
| AraBert | 0.475 | 0.734 | 0.848 | 0.615 |
| AraVec (Wikipedia – SG) | 0.538 | 0.728 | 0.81 | 0.584 |
| AraVec (Wikipedia – CBOW) | 0.411 | 0.646 | 0.772 | 0.518 |

As expected, a decline in the performance of all models was observed, specially with respect to the recall@1 metric. However, the performances of the E5-Large and BGE models were particularly impressive, matching the results of the original experiment at k=5. The experiment involving the BGE model was conducted multiple times to verify the accuracy of the recall@3 results, as they were higher than those observed in the initial experiment. The reasons for this discrepancy are not immediately clear. These initial results seem to indicate that the E5 models as well as the BGE model, are quite resilient to dialect shifts, especially with higher values of k.

Having said that, the authors acknowledge that this experiment was conducted on a very small dataset and used only one dialect. In the future, we aim to explore this area further using larger datasets and a wider variety of dialects.

*4.1.3. Experiment 3: Investigating the impact of using a disambiguating questions on retrieval results*

As mentioned in Section 3.1.2 and shown in Table 1, some questions in the ARCD dataset can only be understood in the context of the preceding question. Section 3.1.2 also detailed how these questions were automatically disambiguated using GPT-3.5 Turbo and subsequently reviewed manually, with both versions being retained. This section presents the results of repeating the experiment described in Section 4.1.1, with both versions of the disambiguated questions. The purpose of this step was to evaluate the performance of the embedding models independently of the ambiguity issue. The outcomes are displayed in Tables 6 and 7, respectively. Values that went down or did not change are marked. All other values went up.

Table 6. Recall @K (k=1, k=3, and k=5) and MRR for the ARCD Dataset with GPT-3.5 auto disambiguation

| Embeddings Name | k=1 | k=3 | k=5 | MRR |
| --- | --- | --- | --- | --- |
| E5-Large | **0.684** ↓ | **0.913** | **0.942** | **0.796** |
| BGE | 0.624 ↓ | 0.885 | 0.922 | 0.752 |
| E5-Small | 0.632 ↔ | 0.875 | 0.915 | 0.752 |
| OpenAI | 0.522 | 0.767 | 0.818 | 0.644 |
| AraVec (Wikipedia – SG) | 0.367 | 0.575 | 0.655 | 0.479 |
| AraBert | 0.343 | 0.556 | 0.626 | 0.454 |
| AraVec (Wikipedia – CBOW) | 0.249 | 0.407 | 0.484 | 0.336 |

Table 7. Recall @K (k=1, k=3, and k=5) and **MRR** for the ARCD Dataset with manually edited disambiguation

| Embeddings Name | k=1 | k=3 | k=5 | MRR |
| --- | --- | --- | --- | --- |
| E5-Large | **0.719** | **0.938** | **0.963** | **0.826** |
| E5-Small | 0.665 | 0.903 | 0.934 | 0.781 |
| BGE | 0.644 | 0.91 | 0.943 | 0.774 |
| OpenAI | 0.544 | 0.777 | 0.831 | 0.662 |
| AraVec (Wikipedia – SG) | 0.389 | 0.597 | 0.664 | 0.499 |
| AraBert | 0.348 | 0.562 | 0.632 | 0.459 |
| AraVec (Wikipedia – CBOW) | 0.255 | 0.414 | 0.493 | 0.344 |

As can be seen from the results, most models perform better after question disambiguation, even with the crude automatic version presented. The presented results also show the E5 models and the BGE model as top performers, with recall@5 results in the nineties.

*4.2. Generator related experiments*

To conclude the exploration of Retrieval-Augmented Generation (RAG) application in Arabic, the final phase was to evaluate various Large Language Models (LLMs) as generators thus completing the pipeline. Since the E5-Large model consistently outperformed all other models on the two used datasets, it was the one used for query and document embedding in the retrieval stage. After the retrieval step was completed, the top 5 returned documents were given to the generators listed in section 3.4 along with the question for which an answer is desired. For the Ar_EduText dataset, experiments were carried out using GPT3.5 Turbo, JAIS 7B quantized, LLama3, Mistral, and Mixtral, while for the ARCD dataset, only the last three open-source LLMs were employed. Each of the open-source LLMs, generated superfluous text in which the answer to the query was often embedded. To apply the chosen metrics as accurately as possible, post-processing functions were written for each LLM after observing the pattern of its generated outputs. All post processing functions can be found in the project's GitHub repository[7]. The results of this experiment are presented in Table 8 and 9, and sample output from the ARCD dataset is presented in Fig. 3

Table 8. Performance of Various LLMs as Generators on the Ar_EduText dataset (all scores are averages).

|  | F1 Score | Bleu Score | Cosine Similarity |
| --- | --- | --- | --- |
| GPT3.5 Turbo | **0.59** | **0.33** | **0.95** |
| Llama3 (llama_70b) | 0.43 | 0.2 | 0.91 |
| Mistral (Mistral_7b) | 0.26 | 0.08 | 0.89 |
| Mixtral | 0.45 | 0.13 | 0.84 |
| JAIS (jais_7b_quantized) | 0.37 | 0.17 | 0.72 |

Table 9. Performance of Various LLMs as Generators on the ARCD dataset (all scores are averages).

|  | F1 Score | Bleu Score | Cosine Similarity |
| --- | --- | --- | --- |
| Mistral (Mistral_7b) | **0.22** | 0.07 | **0.83** |
| Llama3 (llama_70b) | 0.18 | **0.08** | 0.82 |
| Mixtral | **0.22** | 0.04 | 0.79 |

| Question | |
| --- | --- |
| | ما هو اسم الباراسيتامول بالانجليزية؟ |
| Gold Standard Answer | Paracetamol |
| Mistral (Mistral_7b) Answer | Paracetamol or Acetaminophen |
| Llama3 (llama_70b) Answer | اسم الباراسيتامول بالإنجليزية هو Acetaminophen. ويسمى أيضاً Paracetamol في بعض الدول. ويستخدم كمسكن وخافض للحرارة. |
| Mixtral Answer | اسم الباراسيتامول باللغة الإنجليزية هو ويسمى أيضاً بالخَلَنْجُول وهو مسكن وخافض للحرارة واسع الاستخدام اسم الباراسيتامول باللغة الإنجليزية هو ويسمى أيضاً بالخَلَنْجُول |
| Question | |
| | ما هو الباراسيتامول؟ |
| Gold Standard Answer | هو مسكن وخافض للحرارة واسع الاستخدام. |
| Mistral (Mistral_7b) Answer | The barasitamol is a medication that was first created by Harmon Northrop Morse in 1878 through the reduction of barbituric acid with mercury cyanide in an acetic acid solution. However, it was not used in medical treatment until 1893, when it was discovered in urine samples of patients who had consumed it unintentionally. In 1899, it was found that barasitamol was derived from aspirin.[ English text …] |
| Llama3 (llama_70b) Answer | الباراسيتامول هو مسكن وخافض للحرارة واسع الاستخدام. ويصنف ضمن مسكنات الألم من النوع 2. ويؤثر على نفس مستقبلات المورفين، وهو منافس على المستقبلات المورفينية. |
| Mixtral Answer | الباراسيتامول هو مسكن وخافض للحرارة واسع الاستخدام، اكتشف لأول مرة في سنة 1878 بواسطة هارمون نورتروب مورسه، ويصنف ضمن مسكنات الألم من النوع 2 الباراسيتامول هو مسكن وخافض للحرارة واسع الاستخدام، اكتشف لأول مرة في سنة 1878 بواسطة هارمون نورتروب مورسه، ويؤثر على نفس مستقبلات المورفين |

Fig. 3. Sample of ARCD Questions and Answers Generated by the used LLMs

---

[7] https://github.com/SElBeltagy/ArRagExperiments

On the Ar_EduText dataset, GPT3.5 turbo performed best, followed closely in terms of cosine similarity between gold standard answers and generated answers, by Llama 3. The quantized version of JAIS did not do so well, appearing at the bottom of the list. On the ARCD dataset, the number one performer was Mistral followed very closely by Llama3. As can be seen by the sample answers presented in Fig. 3, better prompting might be able to improve the results.

## 5. Conclusion and Future work

This research started out with the expectation that existing embedding models and Large Language Models (LLMs) would face significant challenges in processing Arabic text effectively. This assumption stems from the unique linguistic features of Arabic, including its rich morphology, complex syntactical structures, and diverse dialect base which often pose difficulties for standard NLP models developed predominantly for English. However, the empirical evaluation presented in this work contradicts these initial expectations, demonstrating a notable degree of proficiency and applicability of these models to Arabic texts. In terms of semantic embedding models, it was observed the E5-large model as well as the BGE model show great potential for use with Arabic retrievers and for semantic representation in general.

Furthermore, experiments carried out on various open source LLMs, show that Llama3 and Mistral have great potential as Arabic generators. This finding is instrumental, suggesting that existing open source LLMs can be leveraged to contribute significantly to the development of effective Arabic NLP applications. Future work should investigate the role of prompt engineering and fine tuning to increase the performance of these LLMs even further.

Despite these encouraging outcomes, the authors acknowledge the necessity for broader research to fully investigate the potential of these models. Specifically, future work should explore the application of presented models to a wider array of Arabic dialects and on bigger datasets as well as explore more complicated RAG pipelines.

## Appendix A. Prompt Used to automatically disambiguate questions

```
"""
    Question 1: {q1}
    Question 2: {q2}
    You are an expert who understands Arabic fluently. Given these two questions, your task
is to rephrase the second question only if it contains ambiguities that might confuse
someone without context from the first question. An ambiguity might be a vague reference or
unclear term that cannot be understood without additional context. Do not modify question 2
if it is clear and understandable on its own. Always maintain the response in Arabic.
    Example 1:
    Question 1: من هو حمزة بن عبد المطلب؟
    Question 2: بما وصفه رسول الله؟
    Correct modification: بما وصف رسول الله حمزة بن عبد المطلب ؟
    Example 2:
    Question 1: كم يبلغ ارتفاع مكة عن سطح البحر؟
    Question 2: اين تقع مكة؟
    Correct modification: اين تقع مكة؟
    Remember, modifications are only needed if they clarify ambiguities directly related to
the context provided by question 1. Any name or specific noun is not considered ambiguous.
"""
```